\documentclass[conference]{IEEEtran}
\IEEEoverridecommandlockouts

\usepackage{cite}
\usepackage{amsmath,amssymb,amsfonts}
\usepackage{algorithmic}
\usepackage{graphicx}
\usepackage{textcomp}
\usepackage{xcolor}

\usepackage[english]{babel}
\usepackage{bm}
\usepackage{multirow}
\usepackage{multicol}
\usepackage{graphicx}
\usepackage{lipsum}

\def\BibTeX{{\rm B\kern-.05em{\sc i\kern-.025em b}\kern-.08em
    T\kern-.1667em\lower.7ex\hbox{E}\kern-.125emX}}
\begin{document}

\title{Retrieval-Augmented Speech Recognition Approach for Domain Challenges
    \thanks{This work is partially supported by JSPS KAKENHI No.24K15004.}
}

\author{\IEEEauthorblockN{Peng Shen, Xugang Lu, Hisashi Kawai}
    \IEEEauthorblockA{National Institute of Information and Communications Technology (NICT), Japan \\
        peng.shen@nict.go.jp}
}

\maketitle

\begin{abstract}
    Speech recognition systems often face challenges due to domain mismatch, particularly in real-world applications where domain-specific data is unavailable because of data accessibility and confidentiality constraints. Inspired by Retrieval-Augmented Generation (RAG) techniques for large language models (LLMs), this paper introduces a LLM-based retrieval-augmented speech recognition method that incorporates domain-specific textual data at the inference stage to enhance recognition performance. Rather than relying on domain-specific textual data during the training phase, our model is trained to learn how to utilize textual information provided in prompts for LLM decoder to improve speech recognition performance. Benefiting from the advantages of the RAG retrieval mechanism, our approach efficiently accesses locally available domain-specific documents, ensuring a convenient and effective process for solving domain mismatch problems. Experiments conducted on the CSJ database demonstrate that the proposed method significantly improves speech recognition accuracy and achieves state-of-the-art results on the CSJ dataset, even without relying on the full training data.
\end{abstract}


\section{Introduction}

Automatic speech recognition (ASR) techniques have improved significantly due to advancements in system architecture and optimization algorithms \cite{Yu2014ASR, jinyuLi2021E2EASR, Chan2016ListenAA, Li2020DevelopingRM}. However, in real-world applications, these systems continue to face challenges, often leading to a decrease in recognition accuracy. A major challenge is domain mismatch, which becomes particularly problematic when model training lacks sufficient domain-specific data due to data accessibility and confidentiality constraints.

Recently, with the successful deployment of large language models (LLMs) \cite{openai2023gpt4, touvron2023llama2openfoundation, bai2023qwen}, LLM-based ASR models have gained widespread attention \cite{chu2023qwenaudio, das2024speechverse, kong2024audioflamingo, li2023prompting, wang2023slm, chen2023xllm, rubenstein2023audiopalmlargelanguagemodel, tang2024salmonngenerichearingabilities,suh2024improvingdomainspecificasrllmgenerated, hu2024largelanguagemodelsefficient}.
LLM-based ASR models typically use a pretrained audio encoder as the speech encoder, for example, HuBERT \cite{hsu2021hubertselfsupervisedspeechrepresentation}, the encoder component of Whisper \cite{radford2022whisper}. This speech encoding can be integrated into LLMs either through cross-attention mechanisms or by concatenating the embeddings of audio and other tokens for further processing.

Unlike traditional smaller language models used in ASR, LLMs typically have a large number of parameters and are trained on massive datasets, such as 2 trillion tokens for LLaMA2 and 15 trillion tokens for LLaMA3. This extensive training equips them with strong language generation and semantic understanding capabilities, which can be beneficial for addressing domain mismatch problems.
Recent studies have explored using LLMs for ASR error correction based on ASR N-best outputs \cite{ma2023generativelargelanguagemodels}.
Other methods include prompt engineering approaches, such as reranking N-best hypotheses of an ASR system with LLaMA or applying deep LLM fusion, where LLMs are integrated into the decoder of an encoder-decoder-based ASR system \cite{li2023promptinglargelanguagemodels}. Previous work has also shown that optimizing both the LLM loss and the traditional ASR connectionist temporal classification (CTC) loss can help address the domain mismatch problem \cite{hono2024integratingpretrainedspeechlanguage}.

Despite being trained on large amounts of data, LLMs still struggle to access the latest information or specific domain content and often suffer from hallucination issues. To address these problems, retrieval-augmented generation (RAG) techniques have been proposed and have recently gained increased attention \cite{NEURIPS2020_6b493230, guu2020realmretrievalaugmentedlanguagemodel}. By retrieving up-to-date data or querying a relevant knowledge base in response to specific queries, RAG provides the LLM with relevant information, effectively mitigating the hallucination issue and improving overall performance.
\begin{figure*}[!ht]
    \centering
    \includegraphics[width=490pt]{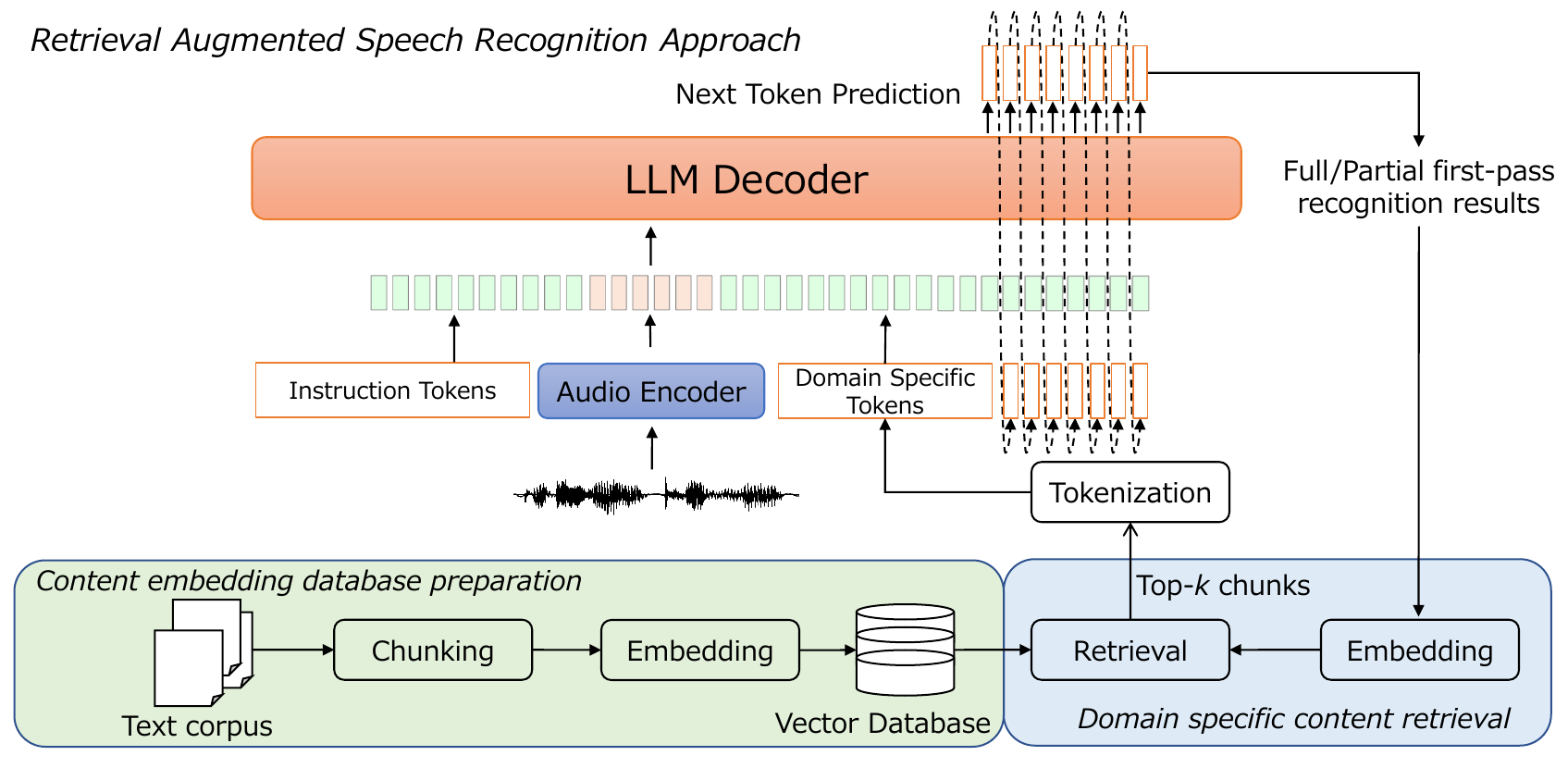}\\
    \caption{Overview of the proposed retrieval augmented speech recognition (RASR) approach.}
    \label{fig.propose_method1}
\end{figure*}

In this paper, inspired by RAG, we integrate retrieval-augmented techniques into an LLM-based speech recognition system to address the challenge of domain mismatch. By introducing an instruction prompt, our method optimizes the model to effectively leverage domain-specific textual content provided in the context, thereby improving speech recognition accuracy.
Rather than requiring the use of domain-related data during the training phase, the proposed method retrieves and utilizes domain-specific data only during the inference stage. This approach leads to significant performance improvements on target domain data while ensuring the protection of sensitive information.
To the best of our knowledge, such approach are still not be applied on ASR tasks. The contributions of this work can be summarized as follows:
\begin{itemize}
    \item We propose a novel approach that leverages domain-specific data through a RAG-based retrieval mechanism during inference and trains the model to effectively utilize domain-specific textual information to improve speech recognition performance.
    \item We provide a detailed investigation to show the importance of the instruction prompt, domain specific content, and the training strategies.
    \item By integrating the retrieval mechanism, our approach efficiently accesses locally domain-specific documents during inference, leading to performance improvements while ensuring the protection of sensitive data.
    \item Experiments on the CSJ dataset demonstrate that the proposed system achieved a 19.6\% relative improvement on the out-of-domain set and state-of-the-art results on the official evaluation set, while utilizing only a fraction of the typical training data.
\end{itemize}


\section{Retrieval Augmented Speech Recognition}
\subsection{Overview of the Proposed Approach}

As shown in Fig. \ref{fig.propose_method1}, to handle domain-specific contextual information, we integrate a RAG-based approach into our LLM-based speech recognition system. The proposed method consists of three key components: content embedding database preparation, domain-specific content retrieval, and LLM-enhanced speech recognition.
Similar to \cite{chu2023qwenaudio, wang2023slm}, the LLM-enhanced speech recognition component is implemented using an LLM decoder paired with an audio encoder. The retrieved contextual information is fed into the LLM decoder as context by combining it with the instruction and audio embeddings, enabling more accurate next-token inference.

Given a speech input $\mathbf{x}$, its corresponding transcription $y$, and domain-specific content $D$, and $I$ as the instruction to guiding the LLM to utlize $D$ for improving prediction. The primary objective is to maximize the probability of the next token $y_{t}$, formulated as:
\begin{equation}
    P(y_t|y_{<t}, I, \text{Encoder}(\mathbf{x}), D),
    \label{eq.one}
\end{equation}
where $I$ instructs the LLM to leverage domain-specific content $D$ to accurately generate the transcription $y$ corresponding to the speech input $\mathbf{x}$.

\subsection{Domain Specific Content Retrieval}
To train the model and perform inference, domain-specific documents need to be retrieved for both training and inference stages. In this work, we use an embedding-based retrieval approach, which consists of two main parts: content embedding database preparation and domain-specific content retrieval.
For the content embedding database, the preparation process involves the following steps:
\begin{enumerate}
    \item \textbf{Document Processing:} We use the transcriptions of the training data as documents for domain-specific retrieval. Preprocessing is applied to the transcriptions of each talk to normalize and concatenate them into a single document.
    \item \textbf{Chunk Splitting:} Each document is divided into smaller chunks, which helps improve retrieval accuracy by providing more granular data.
    \item \textbf{Embedding Generation:} Each chunk is processed through an embedding model, converting the textual data into high-dimensional vectors. These vectors are used for semantic similarity retrieval between the chunks and the query.
    \item \textbf{Storage:} The generated embeddings are stored in a vector database, optimized for quick retrieval during inference stage.
\end{enumerate}

Domain-specific content retrieval is handled differently during the training and inference phases:
During the training phase, retrieval is performed using the transcriptions of the utterance-based speech data.
In the inference phase, queries consist of full or partial recognized results generated by an ASR system.
All retrieval operations are based on cosine similarity, which measures the cosine of the angle between the embeddings of the query and the document chunks.
The top-$k$ chunks (excluding chunks from the same talk) with the highest cosine similarity scores are selected as the domain-specific documents corresponding to the input speech data.
These retrieved document chunks are subsequently integrated into the LLM decoder, alongside the instruction prompt and the representation of the speech.

\subsection{Two-Stage Model Optimization}
\label{sec.twostate.model.optimizaion}

\begin{figure}
    \centering
    \includegraphics[width=250pt]{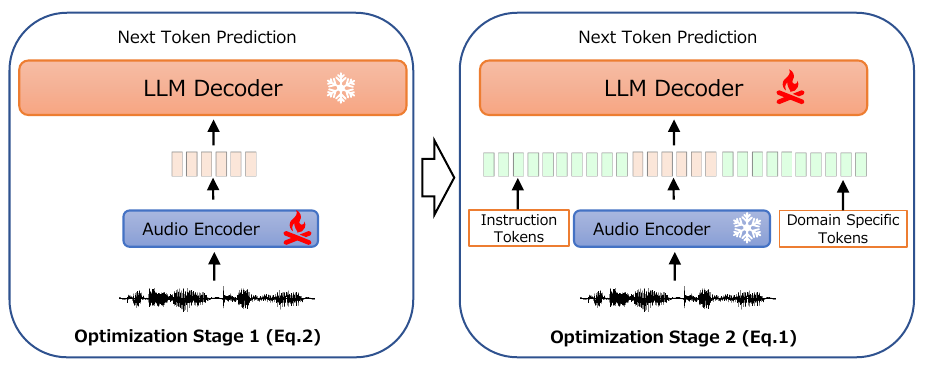}\\ \relax
    \caption{Two-Stage Model Optimization}
    \label{fig.two_stages}
\end{figure}

As illustrated in Fig. \ref{fig.two_stages}, the training of our speech recognition model is structured into two distinct stages, each designed to optimize different components of the system:
\begin{enumerate}
    \item \textbf{Audio Encoder Optimization:}
          In the first stage, we focus solely on training the audio encoder using only $\mathbf{x}$ as input. The objective is to maximize the probability of predicting the next token as follows:
          \begin{equation}
              P(y_t|y_{<t}, \text{Encoder}(\mathbf{x})).
              \label{eq.two}
          \end{equation}
          During this phase, the parameters of the audio encoder are updated, while the LLM decoder parameters remain fixed. This approach ensures that the encoder learns to capture the nuances of the audio signals without being influenced by the decoding process.

    \item \textbf{LLM Decoder Optimization:}
          The second stage of training focuses on the LLM decoder. At this stage, the audio encoder's parameters are frozen, and only the decoder's parameters are updated. As described in Eq. \ref{eq.one}, during this phase, we introduce the instruction prompt and domain-specific content. The instructon prompt \footnote{\textit{Task Instruction: Transcribe the Audio strictly following its content. Use context to verify technical terms and domain-specific vocabulary when uncertain. Ensure the transcription reflects exactly what is spoken, with context aiding in clarifying domain-related ambiguities.}} serves as a guide for the LLM decoder, instructing it on how to effectively incorporate and utilize the domain-specific content retrieved through the RAG process.
\end{enumerate}

\section{Experimental setting and results}
\subsection{Dataset and Evaluation Metrics}

To assess the effectiveness of our proposed speech recognition system, we utilized the Corpus of Spontaneous Japanese (CSJ) for evaluation. The CSJ is a comprehensive dataset consisting of 660 hours of spontaneous Japanese speech, primarily from academic presentations and simulated public speeches, with recordings typically ranging from 10 to 25 minutes in length.
A unique feature of the CSJ corpus is its inclusion of domain-specific texts from fields such as science, engineering, humanities, and social sciences, which are not commonly represented in standard speech corpora. These traits make the CSJ particularly valuable for testing the out-of-domain robustness of ASR systems.

The original evaluation set does not focus on the domain mismatch problem, as the domains are the same as those in the training data. To address this, we excluded talks related to "speech recognition," "language model," and "acoustic model" from the training set. Using the excluded data, we created a new test set (Test) consisting of 20 talks, ensuring that the domain of the test data was not represented in the training set. As a result, our model was trained using only a subset of the available training data, totaling approximately 610 hours.
In addition to the Test dataset, we also used the official evaluation sets as a reference for building our baseline systems. To assess the impact of longer texts, we segmented the speech data into clips no longer than 30 seconds. The character error rate (CER) was used as the evaluation metric.

\subsection{Experimental Setup}

As described in subsection \ref{sec.twostate.model.optimizaion}, the proposed model was trained in two stages. First, the encoder was optimized to enhance acoustic feature extraction, followed by a second step where the LLM decoder was fine-tuned to better utilize domain-specific context for speech recognition. For a further investigation, we also conducted evaluations where the encoder was optimized only in the first stage for 10 and 30 epochs. Additionally, we evaluated the model by optimizing the encoder using both stage 1 (10 epochs) and stage 2 (20 epochs), and optimizing both the encoder and decoder using the settings from stage 1.
The weights of the proposed model were initialized using the whisper-large-v2 model \cite{radford2022whisper} for the audio encoder and the ELYZA-Japanese-Llama-2-7b model \cite{elyzallama2023} for the LLM decoder component.
The whisper-large-v2 model is a 32-layer Transformer architecture that includes two convolutional down-sampling layers \cite{radford2022whisper}. and the ELYZA-Japanese-Llama-2-7b model is also a 32-layer Transformer decoder with a hidden size of 4096, totaling approximately 7 billion parameters \cite{touvron2023llama2openfoundation}.

For the content embedding database preparation, we used the transcriptions of the original training data and set the chunk size to 512 with a 50 overlap. The multilingual-e5-large model \cite{wang2024multilingual} was used to extract embeddings, with an output dimension of 1024. These embeddings were stored in a Chroma vector database, which facilitated fast retrieval operations.
During the inference stage, we evaluated either partial (i.e., the first 30 or 100 characters) or full first-pass recognition results as the query to search for domain-specific documents. The query was first embedded using the multilingual-e5-large model, and cosine similarity was applied to retrieve the top-$k$ chunks (excluding chunks from the same talk of the speech data), where $k$ was set to 2.

The model was trained using the Adam optimizer with a warm restart learning rate scheduler \cite{Loshchilov2016Warmup}, a global mini-batch size of 128, and an initial learning rate of 5e-5, adjusted by a cosine annealing schedule for both two stages.

\subsection{Experimental Results}
\vspace{-0.5em}

\begin{table}[h]
    \centering
    \caption{Comparison of the proposed method with previous SOTA approaches (CER \%). "InD" refers to in-domain; "OoD" refers to out-of-domain.}
    \label{tab.asr_results}
    \begin{tabular}{|l|c|c|}
        \hline
        \textbf{Method}                                                                   & \textbf{Eval1-InD} & \textbf{Test-OoD} \\ \hline
        ESPnet (Conformer) \cite{watanabe2018espnet}                                      & 4.5                & -                 \\ \hline
        ESPnet (Large Transformer) \cite{watanabe2018espnet}                              & 4.9                & -                 \\ \hline
        Whisper large v2 \cite{hono2024integratingpretrainedspeechlanguage}               & 23.7               & -                 \\ \hline
        Whisper large v3 \cite{hono2024integratingpretrainedspeechlanguage}               & 17.2               & -                 \\ \hline
        LLM-based w/ domain adaptation \cite{hono2024integratingpretrainedspeechlanguage} & 5.4                & -                 \\ \hline\hline
        Our baseline (Stage1-10eps)                                                       & 5.1                & 5.5               \\ \hline
        Our baseline (Stage1-30eps)                                                       & 4.6                & 4.6               \\ \hline
        Our proposed (Stage1-10eps + Stage2-20eps)                                        & 4.2                & 3.7               \\ \hline
    \end{tabular}
\end{table}
\vspace{-0.5em}
Table \ref{tab.asr_results} presents a comparison between our proposed method and previous state-of-the-art (SOTA) results on the CSJ dataset. We include two SOTA results referenced from the ESPnet toolkit \cite{watanabe2018espnet}. One model, using a Conformer architecture, demonstrates competitive performance on the Eval1 set with a CER of 4.5\%. Another large transformer-based model achieves a CER of 4.9\%.
Additionally, we list the results from Whisper's large v2 and v3 models, as well as an LLM-based domain adaptation approach evaluated by \cite{hono2024integratingpretrainedspeechlanguage}. Whisper large v2 and v3 models obtained CERs of 23.7\% and 17.2\%, respectively. These results indicate that even models trained on vast datasets, such as Whisper large v3's 5 million hours training data, still suffer from domain mismatch, leading to a significant drop in recognition accuracy. The LLM-based domain adaptation approach, which incorporates domain-specific data for model optimization, showed improved performance.

Our LLM-based baseline implementation, despite being trained on only a subset of the available training data, achieved results comparable to SOTA models, with a CER of 4.6\% on the Eval1 set. Furthermore, by incorporating retrieval-augmented context, the proposed method achieved a CER of 3.7\% on the out-of-domain Test dataset and 4.2\% on the in-domain Eval1 set.
Compared to our baseline system, the proposed method yielded a relative improvement of 19.6\% on the Test set and 8.7\% on the Eval1 in-domain dataset.
This highlights the significant advantage of using domain-specific content for context, as it greatly reduces error rates in domain mismatch ASR tasks.



\subsection{Investigation and Ablation Study}
\label{sec.investigation}
\vspace{-0.5em}

\begin{table}[htbp]
    \centering
    \caption{Investigation on Optimization strategies (CER\%). }
    \label{table.optimization}
    \begin{tabular}{|l|c|c|}
        \hline
        \textbf{Optimization Strategies}        & \textbf{Eval1} & \textbf{Test} \\
        \hline\hline
        S1: enc-stage1-10eps                    & 5.15           & 5.56          \\\hline
        S2: enc-stage1-30eps                    & 4.57           & 4.63          \\\hline
        S3: enc-stage1-10eps + dec-(Eq.2)-20eps & 7.19           & 5.84          \\\hline
        S4: enc-stage1-10eps + enc-(Eq.1)-20eps & 5.07           & 4.22          \\\hline
        S5: enc-stage1-10eps + dec-stage2-20eps & 4.21           & 3.73          \\\hline
    \end{tabular}
\end{table}

To clarify the impact of different settings in the proposed method, we conducted a series of investigations. Firstly, we investigated the optimization strategies for the encoder, decoder, and the two-stage optimization process. The results are presented in Table \ref{table.optimization}.
We began by comparing the performance of optimizing the encoder for 10 epochs (S1) versus 30 epochs (S2). The results showed that extending the training to 30 epochs led to noticeable improvements. Next, for a fair comparison, starting from the model trained for 10 epochs, we further optimized the model by training it for an additional 20 epochs.
We observed that further fine-tuning the decoder component with Eq. \ref{eq.two}, i.e., S3, led to poor results, even when the learning rate was reduced.
Fine-tuning the encoder with Eq. \ref{eq.one}, i.e., S4, improved performance on the Test dataset but only provided marginal improvements on the in-domain Eval1 dataset.
The proposed method, i.e., S5, achieved the best performance on both the Test and Eval1 datasets.

\setlength{\tabcolsep}{7pt}
\begin{table}[h]
    \centering
    \caption{The influence of instruction prompt ($I$), and domain-specific content ($D$). "Rand" refers to randomly selected content, while "30", "100," and "full" represent the use of the first 30 characters, 100 characters, and the full recognition result, respectively. "Trans" refers to using the true label transcription. (CER \%).}
    \label{tab_result_investigation}
    \begin{tabular}{|c|c|c|c|c|c|c|c|c|c|c|} \hline
        \textbf{$I$} & \textbf{$D$:Rand} & \textbf{$D$:30  } & \textbf{$D$:100 } & \textbf{$D$:Full} & \textbf{$D$:Trans} & \textbf{Test} \\ \hline\hline
                     &                   &                   & \checkmark        &                   &                    & 3.97          \\ \hline
                     &                   &                   &                   & \checkmark        &                    & 3.86          \\ \hline
        \checkmark   & \checkmark        &                   &                   &                   &                    & 4.06          \\ \hline
        \checkmark   &                   & \checkmark        &                   &                   &                    & 3.89          \\ \hline
        \checkmark   &                   &                   & \checkmark        &                   &                    & 3.82          \\ \hline
        \checkmark   &                   &                   &                   & \checkmark        &                    & 3.73          \\ \hline
        \checkmark   &                   &                   &                   &                   & \checkmark         & 3.67          \\ \hline
    \end{tabular}
\end{table}

Table \ref{tab_result_investigation} illustrates the investigation into the influence of using the instruction prompt ($I$) and domain-specific content ($D$). The results demonstrated that incorporating the instruction prompt reduced the CER from 3.86\% to 3.73\%. Additionally, domain-specific content retrieved based on first-pass recognition results outperformed randomly selected content, with the CER dropping from 4.06\% to 3.73\%.
Moreover, we observed that longer queries (i.e., using more text in the query) led to better performance. When using the full recognition results as the query, the CER approached that of the transcription-based query, with CER values of 3.73\% and 3.67\%, respectively.

\section{Conclusions}
\vspace{-0.5em}
In this paper, we proposed a novel LLM-based retrieval-augmented speech recognition system that effectively addresses domain-specific challenges by leveraging a RAG-based retrieval mechanism during the inference phase. Our approach eliminates the need for incorporating domain-specific textual data during the training phase, thereby ensuring data privacy and security, while enhancing the performance of the speech recognition system through dynamic utilization of relevant domain-specific content at inference.
We demonstrated the effectiveness of our method through extensive experiments on the CSJ dataset, achieving state-of-the-art results even with only a fraction of the typical training data.

\bibliographystyle{IEEEbib}
\bibliography{mybib4asr,mybib4llmasr}

\end{document}